%% file: emnlp2018.tex
\documentclass[11pt,a4paper]{article}
\usepackage[hyperref]{emnlp2018}
\usepackage{times}
\usepackage{latexsym}
\usepackage{comment}
\usepackage{url}
\usepackage{amsmath}
\usepackage{graphicx}

\usepackage{tabularx}
    \newcolumntype{L}{>{\raggedright\arraybackslash}X}

\DeclareMathOperator*{\argmax}{arg\,max}

\usepackage{url}

\usepackage{stfloats}   

\usepackage{titlesec}
\titlespacing*{\section}{0pt}{0.5\baselineskip}{0.5\baselineskip}

\aclfinalcopy 

\title{Generating More Interesting Responses in Neural Conversation Models with Distributional Constraints}

\author{
Ashutosh Baheti\hspace{1pt}$^{{{\bf 1}}}$, Alan Ritter\hspace{1pt}$^{{{\bf 1}}}$, Jiwei Li\hspace{1pt}$^{{{\bf 2}}}$,  and Bill Dolan\hspace{1pt}$^{{{\bf 3}}}$
\\[0.4cm]
$^1$Ohio State University, OH, USA \\
$^2$Shannon.AI and Renmin University of China, Beijing, China \\
$^3$Microsoft Research, Redmond, WA, USA\\
{\tt  \{baheti.3,ritter.1492\}@osu.edu} \\
{\tt jiweil@stanford.edu} \\
{\tt  billdol@microsoft.com}
}

\date{}

\begin{document}
\maketitle

\parskip 0pt

\begin{abstract}
  Neural conversation models tend to generate safe, generic responses for most inputs.  This is due to the limitations of likelihood-based decoding objectives in generation tasks with diverse outputs, such as conversation. To address this challenge, we propose a simple yet effective approach for incorporating side information in the form of distributional constraints over the generated responses.  We propose two constraints that help generate more content rich responses that are based on a model of syntax and topics \cite{griffiths2005integrating} and semantic similarity \cite{arora2016simple}. We evaluate our approach against a variety of competitive baselines, using both automatic metrics and human judgments, showing that our proposed approach generates responses that are much less generic without sacrificing plausibility. A working demo of our code can be found at \url{https://github.com/abaheti95/DC-NeuralConversation}.
\end{abstract}

\section{Introduction}
\label{sec:intro}

Recent years have seen growing interest in neural generation methods for data-driven conversation.  This approach has the potential to leverage massive conversational datasets on the web to learn {\em open-domain} dialogue agents, without relying on hand-written rules or manual annotation.  Such response generation models could be combined with traditional dialogue systems to enable more natural and adaptive conversation, in addition to new applications such as  predictive response suggestion \cite{kannan2016smart}, however many challenges remain.

A major drawback of neural conversation generation is that it tends to produce too many ``safe" or generic responses, for example:
 {\em ``I don't know"} or {\em ``What are you talking about ?"}.
This is a pervasive problem that has been independently reported by multiple research groups \cite{li2016diversity,serban2016building,li2016deep}.\footnote{\url{https://research.googleblog.com/2015/11/computer-respond-to-this-email.html}}
The effect is due to the use of conditional likelihood as a decoding objective -- maximizing conditional likelihood is a suitable choice for text-to-text generation tasks such as machine translation, where the source and target are semantically equivalent, however, in conversation there are many acceptable ways to respond. Simply choosing most predictable reply often leads to very dull conversation.
 
\begin{figure}
    \centering
    \includegraphics[width=.5\textwidth]{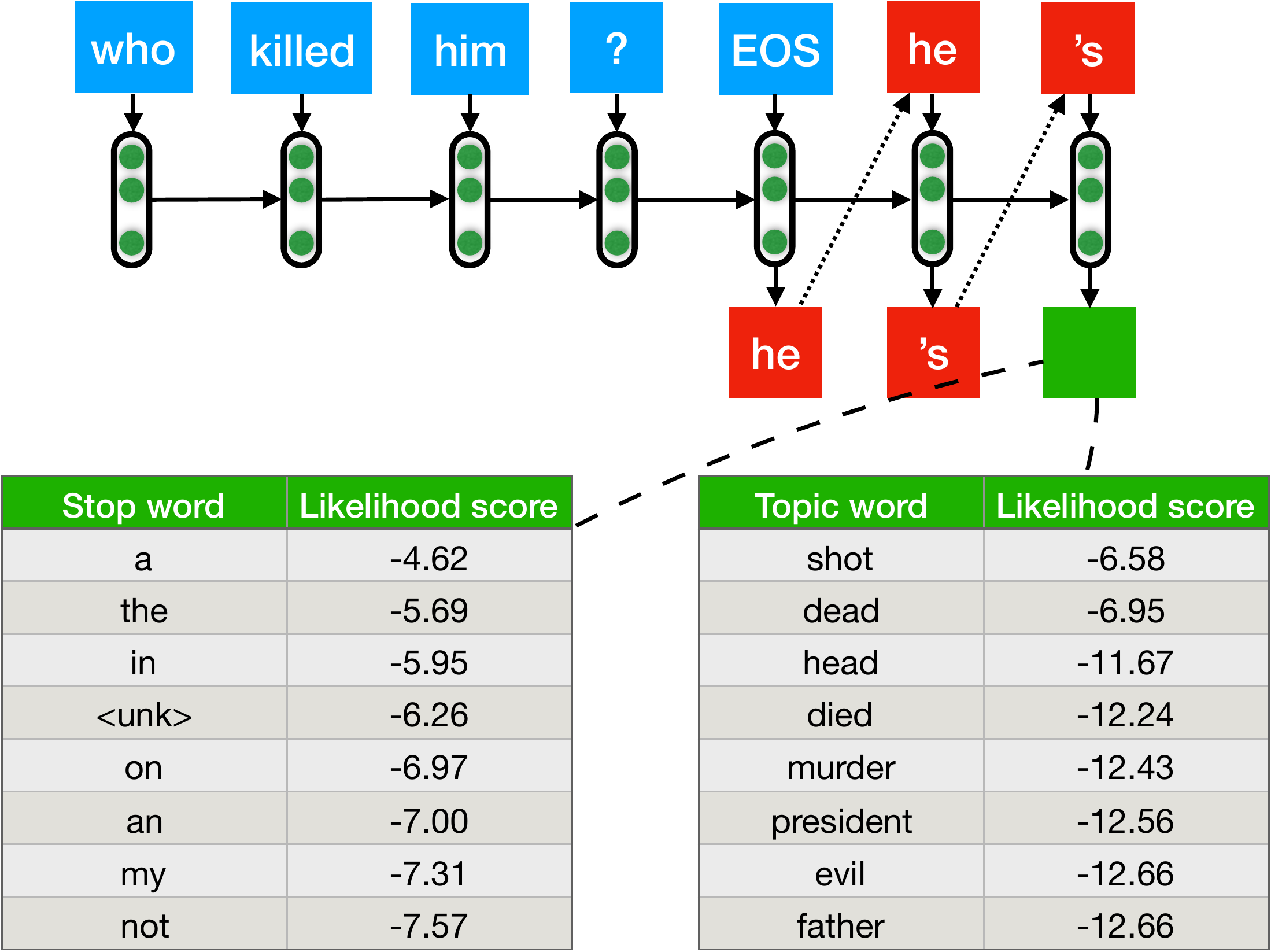}
    \caption{Illustration of the dull response problem in maximum likelihood neural conversation generation using an example from the OpenSubtitles corpus.  Function (stop) words tend to receive higher log probabilities than content (topic) words.  The highest likelihood \href{https://www.ranks.nl/stopwords}{stop words} and topic words in this context are listed.}
    \label{fig:decoder_behavior}
\end{figure}

Figure \ref{fig:decoder_behavior} illustrates the problem with conditional likelihood using an example.  After encoding the source message using a bidirectional LSTM with attention, and fixing the first two words of the response, we show the highest ranked words (according to log-likelihood scores) taken from a list of stop words\footnote{\url{https://www.ranks.nl/stopwords}} in contrast to those selected from a list of topic words.\footnote{The top 10 topic words were taken from each of the 50 topics inferred by an HMM-LDA model (after removing stop words).}  As illustrated in the figure, response generation that is based on maximum likelihood is biased towards stop-words and therefore results in responses that are {\em safe} (likely to be plausible in the context of the input), but also {\em bland} (don't contribute any new information to the conversation).  This motivates the need for augmenting the decoding objective to encourage the use of more content words.
 
To address the {\em dull-response} problem in neural conversation, in this paper, we propose a new decoding objective that flexibly incorporates side-information in the form of distributional constraints.  We explore two constraints, one which encourages the distribution over topics and syntax in the response to match that found in the user's input.  
To estimate these distributions, we leverage the unsupervised model of topics and syntax proposed by Griffiths and Steyvers \shortcite{griffiths2005integrating}.  
The second constraint encourages generated responses to be semantically similar to the user's input; semantic similarity is measured using fixed-dimensional sentence embeddings \cite{arora2016simple}.  

After introducing distributional constraints into the decoding objective, we empirically demonstrate, in an evaluation that is based on human judgments, that our approach generates more content-rich responses when compared with two competitive baselines: Maximum Mutual Information (MMI) \cite{li2016diversity}, in addition to an approach that conditions on topic models as additional context in neural conversation \cite{xing2017topic}.  While encouraging the model to generate less bland responses can be risky, we find that our approach achieves comparable plausibility while introducing significantly more content.

\section{Neural Conversation Generation}
\label{sec:background}
As a starting point for our approach we leverage the Seq2Seq model \cite{sutskever2014sequence,bahdanau2014neural} which has 
been used as a basis for a broad range of recent work on neural conversation \cite{kannan2016smart,li2016diversity,serban2016building,shao2017generating}.
This model consists of two parts, an {\em encoder} and a {\em decoder} both of which are typically stacked LSTM layers. The encoder reads the input sequence and creates a hidden representation. The decoder conditions on this representation, using attention, and generates the response using a neural network language model \cite{bengio2003neural,sutskever2011generating}.

\section{Distributional Topic and Semantic Similarity Constraints}
\label{sec:constraints}
Neural generation models select a response, $\hat{Y}$ by maximizing over a decoding objective, typically using greedy beam search from left to right over partially completed responses, which are scored using the decoder RNN language model. A commonly used decoding objective is the conditional likelihood of the target given the source, $P(Y|X)$:
\begin{eqnarray}
\label{eq:likelihood}
    \hat{Y} & = & \argmax_{Y}\{\log P(Y|X)\} \\
    & = & \argmax_{w_1,\ldots,w_n}\{ \sum_{i=1}^n \log P(w_i|w_1, \ldots w_{i-1},X)\} \nonumber
\end{eqnarray}
As discussed in Section \ref{sec:intro}, models trained to maximize conditional likelihood tend to assign low probability to content words as compared to (more frequent) function words, leading to bland, generic responses most of the time. 
To ameliorate this, we introduce distributional constraints in the form of additional terms in the decoding objective that favor hypotheses containing more content words that are similar to the source in the {\em Topical} and {\em Semantic} sense.


\par For the constraint in the topic domain, we are interested in the topic probability distributions of the source, $X$, and target $Y$, $P(T|X)$ and $P(T|Y)$, where $T$ is a random variable defined over $k$ topics. Then we can modify the decoding objective from Eq \ref{eq:likelihood}:
\begin{equation}
\label{eq:topic_constraint}
\begin{split}
    \hat{Y}^T = \argmax_{Y}\{ &\log P(Y|X) + \\
                            &\alpha \times \Delta(P(T|X),P(T|Y))\}
\end{split}
\raisetag{2.5\normalbaselineskip}
\end{equation}
Here, $\Delta$ is a similarity function between the two probability distributions and $\alpha$ is a tunable hyperparameter to adjust impact of this constraint.

Much recent work has investigated how to encode the semantic meaning of a sentence into a fixed high dimensional embedding space \cite{kiros2015skip, wieting2017revisiting}. Given such an embedding representation of $X$ and $Y$, one can find the semantic similarity between the two and similar to Eq \ref{eq:topic_constraint} we can add a semantic similarity constraint to the likelihood objective as follows:
\begin{equation}
\label{eq:semantic_constraint}
\begin{split}
    \hat{Y}^{Emb} = \argmax_{Y}\{ &\log(P(Y|X)) + \\
                            &\beta \times \Delta(Emb(X),Emb(Y))\}
\end{split}
\raisetag{2.5\normalbaselineskip}
\end{equation}
where, $Emb()$ is a function that maps an utterance to a semantic vector representation, $\Delta$ is a function that computes similarity of the two embeddings and $\beta$ is a tunable parameter.

Both of the constraint terms from Eq \ref{eq:topic_constraint} and Eq \ref{eq:semantic_constraint} are additive in nature and thus can be combined in a straightforward fashion. This formulation allows us to systematically combine information from three different models to produce better responses in terms of topic and semantic relevance. Conceptually, the likelihood term governs the grammatical structure of the response while the topic and semantic constraints drive content selection \cite{nenkova2004evaluating,barzilay2005collective}.

\section{Decoding with Distributional Constraints}
\label{sec:method}
 In Section \ref{sec:constraints}, we defined two constraints (one topic constraint and one semantic) for use in the decoding objective.  Incorporating these constraints during decoding requires that they factorize in a way that is compatible with left-to-right beam search over words in the response. The standard approach to computing posterior distributions in topic models requires a probabilistic inference procedure over the entire source and target.  Furthermore, computing semantic representations can involve the use of complex neural architectures.  Both of these proceedures are difficult to integrate into decoding, because they are computationally expensive and would need to be called repeatedly within the inner loop of the decoder.  Furthermore, when performing left-to-right beam search, as is common practice in neural generation, the complete response is generally not available.
 To address these challenges, we propose using simple additive variants of these methods that factorize over words and which we found to enable efficient decoding without sacrificing performance.



\subsection{Topic Similarity}
Estimating the topic distribution of the source, $P(T|X)$, and response, $P(T|Y)$, is a key step in implementing the topic-similarity constraint. HMM-LDA is a generative model that is able to separate topic and syntax words, by inferring topic distributions in a corpus while flexibly modeling function words. We briefly summarize this model before describing our implementation.

\subsubsection{Syntax-Topics model}
\label{subsec:HMM-LDA}
Griffiths et. al. \shortcite{griffiths2005integrating} suggested an unsupervised generative model that simultaneously labels each word in a document with a syntax ($c$) and topic ($z$) state. They modify the Latent Dirichlet Allocation model to include a syntactic component akin to a Hidden Markov Model (HMM). In LDA, each topic ($z$) is associated with a probability distribution over the vocabulary $\phi^{(z)}$. HMM-LDA adds additional distributions over words for each syntactic class ($c$) as $\phi^{(c)}$.  A special class, $c=0$, is reserved for topics. The transition model between classes $c_{i-1}$ to $c_i$ follows a multinomial distribution distribution $\pi^{(c_{i-1})}$. Each document has an associated distribution over topics $\theta^{(d)}$; each word, $w_j$, in the document has an associated latent topic variable, $z_j$, that is drawn from $\theta^{(d)}$ and $c_j$ is drawn from $\pi^{(c_{j-1})}$. If $c_j=0$, then $w_j$ is drawn from $\phi^{(z_j)}$, otherwise it is drawn from $\phi^{(c_j)}$.  Markov Chain Monte Carlo inference (MCMC) is used to infer values for the hidden topic and syntax variables associated with a given document collection.  To estimate topic and syntax distributions, we performed collapsed Gibbs sampling over our training corpus of conversations, where each conversation is treated as a document.  One sample of the hidden variables was used to estimate model parameters after 2,500 iterations of burn in.  Our code for training the HMM-LDA model is \href{https://github.com/abaheti95/HMM-LDA}{available online}\footnote{\url{https://github.com/abaheti95/HMM-LDA}}.

\subsubsection{Estimating Topic Distributions with HMM-LDA}
To compute distributional topic constraints in neural response generation, we first need an efficient method for estimating topic distributions that factorizes over words, given a point estimate of an HMM-LDA model's parameters.  We would like to estimate topic distributions based on content words contained in a sentence and ignore function words.
HMM-LDA provides us with topic, $\phi^{(z)}$, and syntax, $\phi^{(c)}$, distributions over the vocabulary of words, $w \in V$.  Treating a sentence as a bag-of-words we can estimate its distribution over topics as a sum of topic distributions over all words normalized by sentence length. However, we found this approach does not to work well in practice because it gives equal weight to topic and syntax words. To address this issue, we weighted each word's topic distribution $P(T|w)$ by its probability of being generated by the topic component of the HMM-LDA model (i.e. $P(C=0|w)$). The topic distribution of a sentence, $S$, is estimated as:
\begin{equation}
\label{eq:T_and_C_given_sent}
\begin{split}
    P(T|S) = \frac{1}{Z}\Sigma_{w \in S} P(T|w) P(C=0|w)
\end{split}
\raisetag{1\normalbaselineskip}
\end{equation}
where $Z = \Sigma_{w \in S}P(C=0|w)$ is a normalizing constant that corresponds to the expected number of content words in the sentence. As mentioned earlier, a more accurate estimate of the topic distribution could be obtained using MCMC inference or by applying the forward-backward algorithm. However, these methods are computationally expensive and not well-suited to the decoding framework used in neural generation.

The method described above allows us to efficiently compute the topic distribution of a sentence for use in the topic constraint in Eq \ref{eq:topic_constraint}. For a similarity function, $\Delta$, we simply use the vector dot product, which is closely related to cosine similarity. This formulation has the advantage that it enables memoization during decoding. Another advantage is that it captures the ratio of topic to syntax words due to the weights $P(C=0|w)$.\footnote{Assuming topic distribution of syntax words to be uniform, a sentence with more syntax words will dampen modes in the distribution. Alternately, with less syntax words the overall distribution will be more peaked.} Therefore, the overall constraint has the effect of keeping the syntax-topics ratio in generated hypothesis similar to the source. 


\subsection{Semantic Similarity}
To define the semantic similarity constraint we first encode a semantic representation of the source and target into a fixed dimensional embedding space. There are many sentence embedding methods that could be used, however we want this encoding to be relatively efficient as it will be used many times during beam search. 

Arora et. al. \shortcite{arora2016simple} recently proposed a simple sentence embedding method, which was shown to have competitive performance across a variety of tasks. Their approach uses a weighted average of word embeddings where each word is weighted by $\frac{a}{a + P(w)}$; here, $P(w)$ is the unigram probability and $a$ is a hyperparameter. Such a weighting scheme reduces the impact of frequent words (typically function words) in the overall sentence embedding. Next the first principal component of all the sentence embeddings in the corpus is removed. \cite{arora2016simple} points that the first principal component has high cosine similarity with common function words. Removing this component gives sentence embeddings that encapsulate the semantic meaning of the sentence.  We use this technique in our implementation of $Emb()$ in Eq \ref{eq:semantic_constraint}. For the similarity function, $\Delta$, we use the dot product.  Analogous to the topic constraint described above, this approach to measuring semantic similarity also decomposes over words and works well in the decoding framework. 

\begin{table}[h]
    \centering
    \begin{tabular}{ |c|c| }
         \hline
         Parameter & Value \\ 
         \hline
         \hline
         RNN Type & Bi-LSTM \\ 
         Layers & 4 \\
         Hidden layer dim. & 1000 \\
         Learning rate & 0.1 \\
         max. grad. norm. & 1 \\
         Optimization & Adadelta \\
         Parameter Init & (-0.08, 0.08) (uniform) \\
         \hline
    \end{tabular}
    \caption{Hyperparameter setting for training}
    \label{tab:train_hyperparams}
\end{table}

\begin{table}[h]
    \centering
    \begin{tabular}{ |c|c|c| }
         \hline
         Bucket & \#dialogues & \#test \\ 
         \hline
         \hline
         b1 (3-6 words) & 10994 & 334 \\ 
         b2 (7-15 words) & 15794 & 333 \\
         b3 (16-25 words) & 5167 & 333 \\
         \hline
         total & 31955 & 1000 \\
         \hline
         
    \end{tabular}
    \caption{Test set from Cornell Movie Dialogue Corpus. Column 2 shows the total number of dialogues that we got after all pre-processing and Column 3 shows the number of sampled dialogues in the test set.}
    \label{tab:test_dataset}
\end{table}

\section{Datasets}
\label{sec:datasets}
For training purposes we use OpenSubtitles \cite{tiedemann2009news}, a large corpus of movie subtitles (roughly 60M-70M lines) that is freely available and has been used in a broad range of recent work on data-driven conversation.  OpenSubtitles does not contain speaker annotations on the dialogue turns, so as previously noted when used for learning data-driven conversation models the data is somewhat noisy. Nonetheless, it is possible to create a useful corpus of conversations from this data by assuming each line corresponds to a full speaker turn. Although this assumption is often violated, prior work has successfully trained and evaluated neural conversation models using this corpus. In our experiments we used a preprocessed version of this dataset distributed by Li et. al. \shortcite{li2016diversity}.\footnote{\hyperlink{http://nlp.stanford.edu/data/OpenSubData.tar}{http://nlp.stanford.edu/data/OpenSubData.tar}} The dataset contains large number of two turn dialogues out of which we sampled 23M to use as our training set and 10k as a validation set.

Due to the noisy nature of the OpenSubtitles conversations we do not use them for evaluation. Instead, we leverage the Cornell Movie Dialogue Corpus \cite{Danescu-Niculescu-Mizil+Lee:11a} which is much smaller but contains accurate speaker annotations. We extracted all two turn conversations (source target pair) from this corpus and removed those with less than three and more than 25 words. After this, we divided the remaining conversations into three buckets based on source length. The numbers can be found in Table \ref{tab:test_dataset}. From each bucket we randomly sampled $\approx$333 dialogues for a total of 1000 dialogues in our test set. We evaluate all models on this test set. Since automatic metrics do not correlate with human judgment, we manually tuned the hyperparameters ($\alpha$ and $\beta$) on a small development set (4 dialogues from each bucket to create a small 12 sentence development set; disjoint from test set). We manually inspected the responses generated by the model on the development set for different values of $\alpha$ and $\beta$ and choose those that performed best.

\section{Experimental Conditions and Baselines}
During learning we use the same hyperparameters for all models; these are displayed in Table \ref{tab:train_hyperparams}, and are based on those reported by Li et. al. \shortcite{li2016diversity}.\footnote{OpenNMT is used for training our models \cite{opennmt}.}  We compare our approach with the following baselines:

\noindent
 {\bf MMI}: We re-implemented the MMI-bidi method proposed by Li et. al. \shortcite{li2016diversity}. MMI is a particularly appropriate baseline for comparison, as it encourages responses that have higher relevance to the input in contrast to conditional likelihood, which tends to favor responses with higher unconditional probability.
 MMI-bidi generates $B$ candidates using Beam search on a Seq2Seq model trained to maximize conditional likelihood of the target given the source, $P(Y|X)$, then re-ranks them using a separately trained source given target model, $P(X|Y)$.  Combining both directions in this way has the effect of maximizing mutual information \cite{li2016diversity}. 
    
\input{model_responses}

\noindent
 {\bf TA-Seq2Seq}: Another relevant baseline is the TA-Seq2Seq model of Xing et. al. \shortcite{xing2017topic} that integrates information from a pre-trained topic model into neural response generation using an attention mechanism to condition on relevant topic words.  They evaluate their model on a dataset of Chinese forum posts.  Unfortunately we could not use the code provided by the authors due to data-mismatch (their model makes use of user identities which are not available in the OpenSubtitles corpus). We therefore compare with a re-implementation of their approach in which we modify each source sentence to include a list of the 20 most relevant topic words from HMM-LDA and then train using the same Seq2Seq framework with attention.  This enables the model to condition on the relevant topic words. In addition to incorporating attention over topics, Xing et. al. also introduced an approach to {\em biased generation} - to replicate this we add a constant factor to all topic words during the prediction. 
    
\section{Results and Analysis}
\label{sec:results}
Our proposed decoding objective constraints (topic and semantic) are complementary to the MMI objective, which encourages diversity and relevance to the source input. Therefore, in addition to comparing against the baselines described above, we evaluated three variants of our model: {\bf (1)} maximum conditional likelihood combined with semantic and topic distributional constraints with a beam size of 10 (DC-10) {\bf (2)} The same configuration with MMI-bidi re-ranking using a beam size of 10 DC-MMI10 and {\bf (3)} MMI-bidi re-ranking with a beam size of 200 (DC-MMI200). We test all configurations on the 1000 conversations test set described in Section \ref{sec:datasets} and compare them on automatic metrics and also in a crowdsourced human evaluation.  We do not consider TA-200 (TA-Seq2Seq, Beam=200), DC-200 and MMI-10 for human evaluation as they appear to perform worse than other model variants in automatic metrics and also on our set of development sentences.  Sample responses for all the remaining models are presented in Table \ref{tab:1}.


\begin{table*}[h]
 \centering
  \begin{tabularx}{\textwidth}{ |p{2in}|c|p{1.6cm}|p{1.6cm}|p{0.85cm}|p{0.83cm}|p{0.93cm}| }
 \hline
 Model & Alias & distinct-1 & distinct-2 & BLEU -1 & Avg. length & Stop-word\% \\
 \hline
 \hline
Human responses & human & 2381/0.176 & 7532/0.602 & - & 13.5 &  70.66\\
 \hline
 MMI (Beam=200) & MMI200 & 351/0.058 & 990/0.197 & 12.8 & 6.0 & 84.91\\
 TA-Seq2Seq(Beam=10) & TA-10 & 237/0.036 & 524/0.095 & 12.9 & 6.5 & 79.40\\
 Dist. Const. (Beam=10) & DC-10 & 710/0.097 & 2014/0.320 & 11.0 & 7.3 & \textbf{72.04}\\
 Dist. Const. + MMI (Beam=10) & DC-MMI10 & 732/0.099 & 2098/0.327 & 11.4 & 7.4 & 73.87\\
 Dist. Const. + MMI (Beam=200) & DC-MMI200 & 850/\textbf{0.116} & 2946/\textbf{0.465} & 11.6 & 7.3 & 72.25\\
 \hline
 \end{tabularx}
 \caption{Automatic metrics evaluation. The $3^{rd}$ and $4^{th}$ columns show the ratio of types to tokens for unigrams and bigrams respectively. $7^{th}$ Column shows the $\%$ of stop-words generated by the models in their responses.}
 \label{tab:automatic_results}
\end{table*}

\begin{table}[t]
 \centering
  \begin{tabular}{ |l|c|c|c| }
 \hline
 Model Alias & No(\%) & Unsure(\%) & Yes(\%)\\
 \hline
 \hline
 \multicolumn{4}{|c|}{Plausible?} \\
 \hline
 human & 19.807 & 23.448 & 56.745\\
 \hline
 MMI200 & 27.623 & 26.445 & 45.931\\
 TA-10 & \textbf{26.981} & 26.874 & \textbf{46.146}\\
 DC-MMI200 & 30.835 & 24.41 & 44.754\\ \hline
 \multicolumn{4}{|c|}{Content Richness?} \\
 \hline
 human & 16.488 & 19.914 &  63.597 \\
 \hline
 MMI200 & 23.662 & 32.976 & 43.362\\
 TA-10 & 31.799 & 30.086 & 38.116 \\
 DC-MMI200 & \textbf{20.021} & 26.660 & \textbf{53.319}\\
 \hline
 \end{tabular}
 \caption{Human judgments for Plausibility of the different models. Each numerical cell contains a percentage value corresponding to its row truncated to 2 decimal precision.}
 \label{tab:mturk_results}
\end{table}

\begin{table}[t]
 \centering
  \begin{tabular}{ |l|c|c|c| }
 \hline
 Model Alias & No (\%) & Unsure (\%) & Yes (\%)\\
 \hline
 \hline
 \multicolumn{4}{|c|}{Plausible?} \\
 \hline
 DC-10 & 36.617 & 27.944 & 35.439\\
 DC-MMI10 & 33.619 & 28.694 & 37.687\\
 DC-MMI200 & \textbf{30.835} & 24.41 & \textbf{44.754}\\ \hline
 \multicolumn{4}{|c|}{Content Richness?} \\
 \hline
 DC-10 & 19.272 & 26.017 & 54.711\\
 DC-MMI10 & \textbf{18.844} & 26.231 & \textbf{54.925}\\
 DC-MMI200 & 20.021 & 26.660 & 53.319\\
 \hline
 \end{tabular}
 \caption{Comparing the model variation by reducing beam size to 10 and also comparing decoder constraints without MMI reranking}
 \label{tab:model_variation_results}
\end{table}

\subsection{Automatic Metrics}
\label{subsec:auto_results}
Following Li et. al. \shortcite{li2016diversity}, we report distinct-1 and distinct-2, which measure the diversity of responses. 
These are the ratios of types to tokens for unigrams and bigrams, respectively.
We also report BLEU-1 scores following previous work, however it should be noted that BLEU-1 is not generally accepted to correlate with human judgments in conversation generation tasks \cite{liu-EtAl:2016:EMNLP20163} as there are many acceptable ways to reply to an input which may not match a reference response. Lastly, we compare the percentage of \href{https://www.ranks.nl/stopwords}{stop-words}\footnote{Long Stopword List from \url{https://www.ranks.nl/stopwords}. We appended punctuations to this list.} of the responses generated by each model (smaller values, that are closer to the distribution of human conversations are preferred). The automatic evaluation is presented in Table \ref{tab:automatic_results}.
\par For brevity we define aliases for each system in the $2^{nd}$ column of Table \ref{tab:automatic_results} which are used in subsequent discussion.  The human responses are diverse and also generally longer than automatically generated responses.  MMI200 has higher diversity than TA-Seq2Seq in terms of distinct-1 and distinct-2. This illustrates the importance of re-ranking using MMI. Our approach produces almost twice as many distinct unigrams and bigrams.  We also observe MMI200 and TA-Seq2Seq achieve higher BLEU scores than our models, however this is not surprising since our models are designed to generate more interesting responses containing rarer content words that are less likely to appear in reference responses. As expected we observe that MMI200 and TA-10 have a higher percentage of stop-words than human responses.  According to the human evaluation discussed in Section \ref{subsec:human_results}, these models were also found to have lower content richness.

\subsection{Human Evaluation}
\label{subsec:human_results}
We conducted a survey on the crowd-sourcing platform, Amazon Mechanical Turk. Every model response is scored on 2 categories: 1) \textbf{P}lausibility - {\em is the response plausible for the given source?} and 2) \textbf{C}ontent Richness - {\em does the response add new information to the conversation?} 
We asked the evaluators to respond on a 5-point scale to the questions above (Strongly Agree, Agree, Unsure, Disagree, Strongly Disagree).  These were later collapsed to 3 categories (Agree, Unsure, Disagree).
The results for plausibility and content richness of our model in addition to the MMI and TA-Seq2Seq baselines and human responses are presented in Table \ref{tab:mturk_results}. 

We observe that MMI200 and TA-10 models achieve slightly better plausibility
scores since they tend to generate safe, dull responses. However, we find that when using a beam size of 200 and MMI re-ranking, our approach which incorporates distributional constraints, DC-MMI200, achieves competitive plausibility, while achieving significantly higher content richness.

\subsubsection{Statistical Significance of Results}
To verify the statistical significance of our findings, we conducted a pairwise bootstrap test \cite{efron1994introduction,berg2012empirical} comparing the difference between percentage of {\em Agree} annotations (Yes column in the Table \ref{tab:mturk_results}). We computed p-values for each pair of models: MMI200 vs DC-MMI200 and TA vs DC-MMI200. For plausibility, we did not find a significant difference in either comparison (p-value $\approx 0.25$) while for content richness, both differences were found to be significant (p-value \textless $10^{-4}$). To summarize: our model significantly beats both baselines in terms of content richness while the difference in plausibility was not found to be statistically significant.

\subsubsection{Pairwise Evaluation of Interestingness}
To further validate our claims we also did a side by side comparison study between MMI200 and DC-MMI200. For every test case, we showed Mechanical Turk workers the source sentence along with responses generated by both systems and asked them select which is more interesting.  We observe that in $56\%$ out of 1000 cases, DC-MMI200 was rated as the more interesting response. The result is statistically significant with p-value \textless $4 \times 10^{-4}$ (using an exact binomial test).

\subsection{Model Variations}
To see the effectiveness of our decoding constraints separately, we compare the best performing DC-MMI200 model with DC-10 and DC-MMI10, both of which use a beam size of 10 -- DC-10 does not include MMI reranking. The results of Mechanical Turk evaluation, following the approach described in Section \ref{subsec:human_results}, are presented in Table \ref{tab:model_variation_results}. We observe that with a beam size of 10 our model is able to generate content rich responses, but suffers in terms of plausibility. The values in the table suggests the decoding constraints defined in this work successfully inject content words into candidate hypotheses and that MMI is able to effectively choose plausible candidates. In the case of DC-10 and DC-MMI10, both models generate the same candidates, but MMI is able to re-rank the results and thus improves plausibility. 

\section{Related Work}

Conversational agents primarily fall into two categories: task oriented dialogue systems \cite{williams2013dialog,wen-EtAl:2015:EMNLP} and chatbots \cite{weizenbaum1966eliza}, although there have been some efforts to integrate the two \cite{dodge2015evaluating,ijcai2017-589}.
Some of the earliest work on data-driven chatbots \cite{ritter2011data} explored the use of phrase-based Statistical Machine Translation (SMT) on large numbers of conversations gathered from Twitter \cite{ritter2010unsupervised}.  Subsequent progress on the use of neural networks in machine translation inspired the use of Sequence-to-Sequence (Seq2Seq) models for data-driven response generation \cite{shang2015neural, sordoni2015neural, li2016diversity}. 

Our approach, which incorporates distributional constraints into the decoding objective, is related to prior work on posterior regularization \cite{mann2008generalized,ganchev2010posterior,zhu2014bayesian}. Posterior regularization introduces similar distributional constraints on expectations computed over unlabeled data using a model's parameters.  These are typically added to the learning objective for semi-supervised scenarios where available labeled data is limited.  In contrast, our approach introduces distributional constraints into the decoding objective as a way to combine neural conversation models trained on large quantities of conversational data with separately trained models of topics and semantic similarity that can drive content selection.

There are numerous examples of related work on improving neural conversation models. Shao et. al. \shortcite{shao2017generating} 
introduced a stochastic approach to beam search that does segment-by-segment reranking to promote diversity. Zhang et. al. \shortcite{zhang2018personalizing} develop models which converse while assuming a persona defined by a short description of attributes. 
Wang et. al. \shortcite{wang2017steering} suggested decoding methods that influence the style and topic of the generated response. Bosselutet al. \shortcite{bosselut2018discourse} develop discourse-aware rewards with reinforcement learning (RL) to  generate long and coherent texts.
Li et. al. \shortcite{li2016deep} applied deep reinforcement learning to dialogue generation to maximize long-term reward of the conversation, as opposed to directly maximizing likelihood of the response.  This line of work was further extended with adversarial learning \cite{li2017adversarial} that rewards generated conversations that are indistinguishable from real conversations in the data.
Lewis et. al. \shortcite{lewis-EtAl:2017:EMNLP2017} applied reinforcement learning with dialogue rollouts to generate replies that maximize expected reward, while learning to generate responses from a crowdsourced dataset of negotiation dialogues.  Choi et. al. \shortcite{choi18} used crowd-workers to gather a corpus of 100K information-seeking QA dialogues that are answerable using text spans from Wikipedia. 
Niu and Bansal \shortcite{niu2018polite} designed a number of weakly-supervised models that generate polite, neutral or rude responses.  Their fusion model combines a language model trained on polite utterances with the decoder. In the second method they prepend the utterance with a politeness label and scale its embedding to vary politeness. The third model is Polite-RL which assigns a reward based on a politeness classifier. 
Gimpel et. al. \shortcite{gimpel2013systematic} explored methods for increasing the diversity of N-best lists in machine translation by introducing a pairwise dissimilarity function. Similar ideas have been explored in the context of neural generation models. \cite{vijayakumar2016diverse, li2016mutual,li2016simple}

\par Following previous work we evaluated our approach using a combination of automatic metrics and human judgments.  Some recent work has explored the possibility of adversarial evaluation of neural conversation models \cite{lowe2017towards,li2017adversarial}.

\section{Conclusions}
We presented an approach to generate more interesting responses in neural conversation models by incorporating side information in the form of distributional constraints. When using maximum likelihood decoding objectives, neural conversation models tend to generate safe responses, such as ``I don't know" for most inputs.  Our proposed approach provides a flexible method of incorporating a broad range of distributional constraints into the decoding objective.  We proposed and empirically evaluated two constraints that factorize over words, and therefore naturally fit into the commonly used left-to-right beam search decoding framework.  The first encourages the use of more relevant topic words in the response the second encourages semantic similarity between the source and target. We empirically demonstrated, through human evaluation, that when taken together these constraints lead to responses that contribute significantly more information to the conversation, while maintaining plausibility in the context of the input.

\section*{Acknowledgments}
We thank the anonymous reviewers for their valuable feedback. 
This material is based upon work supported by the National Science Foundation under Grant No.\ IIS-1464128.


\bibliography{emnlp2018}
\bibliographystyle{acl_natbib_nourl}

\end{document}

%% file: model_responses.tex
\begin{table*}[ht]
  \centering
  \begin{tabularx}{\linewidth}{p{1.2in}|L|L|L|L}
    \hline
    \textbf{Source} & \textbf{Target (ground truth)} & \textbf{MMI} & \textbf{Approximate TA-Seq2Seq} & \textbf{Distributional Constraints} \\ \hline
    where does he work ? & 
    in there , sir . & 
    does he know where he is ? & 
    what kind of place is this ? & 
    he has work to do at the university \\ 
    \hline
    here 's your jacket ! & 
    uh , thanks ... & 
    i don 't want it ! & 
    what kind of suit is this ? & 
    you can stick your shirt in it ! \\ 
    \hline
    what 's so damn funny ? & 
    been to any good \textless unk\textgreater lately ? &
    what are you laughing at ? & 
    what 's the matter with you ? & 
    it 's sort of a funny thing you know ? \\ 
    \hline
    well , what exactly does our platoon do ? serve \textless unk\textgreater ? process paperwork ? & 
    our platoon has the best assignment of all . we 're the first into battle ! & 
    well as a matter of fact & 
    the best thing to do ? & 
    well the war is over but we don 't have time for this shit \\ 
    \hline
    left us here to rot . that 's what they 've done . heroes of the newspapers ! & 
    all right , all right . keep quiet . & 
    they 'r e all dead ! & 
    it 's time to fight ! & 
    we are brothers and kings of war ! \\ 
    \hline
    that 's it . you 're not setting foot off this ship until we return home . & 
    but i haven 't even shown you my biology project ... & 
    we 'r e not going anywhere & 
    i hope you don 't mind & 
    we have orders that the ship will return to earth immediately \\ 
    \hline
    i 'm han solo , captain of this vessel . who 's in charge then ? & 
    i 'm ben kenobi. luke starkiller here is leading our expedition . & 
    i don 't know sir sir & 
    you 'r e a coward ! & 
    i am captain kirk commander of the ship \\ 
    \hline
    her grandmother said she 'd been threatening to run away . and i found the car at the miami bus terminal . & 
    you don 't think sam could 've put it there ? & 
    but she didn 't tell me & 
    that 's one hell of a job & 
    it was parked in a car crash near the road \\ 
    \hline
  \end{tabularx}
  \caption{Sample responses of all the models on the dev set}
  \label{tab:1}
\end{table*}